\documentclass{article}

\usepackage{arxiv}
\usepackage{hyperref}
\usepackage[utf8]{inputenc} 
\usepackage[T1]{fontenc}    
\usepackage{hyperref}       
\usepackage{url}            
\usepackage{booktabs}       
\usepackage{amsfonts}       
\usepackage{nicefrac}       
\usepackage{microtype}      
\usepackage{graphicx}
\usepackage{tabularx}
\usepackage{multirow}
\usepackage{parskip}
\usepackage{float}
\usepackage{subcaption}
\usepackage{graphicx}
\usepackage{enumitem}
\usepackage{cite}
\title{MLRM: A Multiple Linear Regression based Model for Average Temperature Prediction of A Day}
\date{}  

\author{ \href{https://orcid.org/0000-0003-3746-6034}{\includegraphics[scale=0.06]{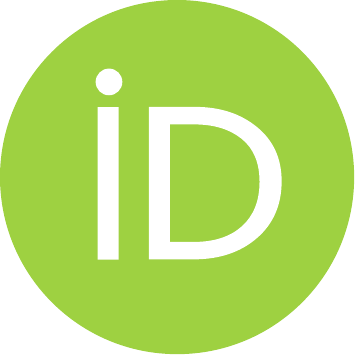}\hspace{1mm}Ishu Gupta*}
	\\
	Cloud Computing Research Center\\
	Department of Computer Science and Engineering\\ 
	National Sun Yat-sen University\\
	Kaohsiung, Taiwan\\
	\texttt{ishugupta23@gmail.com} \\
	\And
	Harsh Mittal \\
	Department of Computer Applications\\
	National Institute of Technology\\
	Kurukshetra, India \\
	136119\\
	\texttt{harsh\_51710010@nitkkr.ac.in}
	\And
	Deepak Rikhari\\
	Department of Computer Applications\\
	National Institute of Technology\\
	Kurukshetra, India \\
	136119\\
	\texttt{deepak\_51710013@nitkkr.ac.in}
	\And
	\href{https://orcid.org/0000-0002-8053-5050}{\includegraphics[scale=0.06]{orcid.pdf}\hspace{1mm}Ashutosh Kumar Singh} \\
	Department of Computer Applications\\
	National Institute of Technology\\
	Kurukshetra, India \\
	136119\\
	\texttt{ashutosh@nitkkr.ac.in} \\
}

\renewcommand{\shorttitle}{\textit{MLRM: A Multiple Linear Regression based Model for Average Temperature Prediction}}

\hypersetup{
pdftitle={A template for the arxiv style},
pdfsubject={q-bio.NC, q-bio.QM},
pdfauthor={David S.~Hippocampus, Elias D.~Striatum},
pdfkeywords={First keyword, Second keyword, More},
}

\begin{document}

\maketitle
\begin{abstract} 
Weather is a phenomenon that affects everything and everyone around us on a daily basis. Weather prediction has been an important point of study for decades as researchers have tried to predict the weather and climatic changes using traditional meteorological techniques. With the advent of modern technologies and computing power, we can do so with the help of machine learning techniques. We aim to predict the weather of an area using past meteorological data and features using the Multiple Linear Regression Model. The performance of the model is evaluated and a conclusion is drawn. The model is successfully able to predict the average temperature of a day with an error of 2.8 degrees Celsius.
\end{abstract}

\section{Introduction}
Weather prediction is the science of predicting the conditions of the atmosphere at a particular time and place \cite{Saxena,GUIM-SMD,MACI}. This prediction has been attempted formally and informally for decades using quantitative data and measurement tools \cite{forecasting,MLPAM,Kaur2017}. Traditionally, meteorologists use various tools for various purposes \cite{SELI,Godha,Confidentiality,Tiwari}. Some of these tools include thermometers for temperature measurement, barometers for air pressure measurement, anemometers for wind speed measurement, etc. \cite{Holistic,Kesharwani,CSA,BatraGarima}.  With the advancement of new tools and technologies and an increase in data and computing power, we can use the data collected by these devices to predict the weather using modern techniques which include statistical techniques and machine learning \cite{Weather,JCOMSS,IDS,kaur2017comparative}. Weather prediction can be called a form of data mining that is concerned with finding hidden patterns inside largely available meteorological data \cite{Kohail,Sharma,Arora,Nishad}.

An accurate weather prediction is advantageous for a significant amount of the population. In fact, every person and everything happening around us is affected by the current weather condition \cite{DT-ILIS,Ayushi,Sloni,singh2020survey}. Some of the areas include disaster management farming, navigation for ships and airplanes, operation of hydro-power plants, sports and entertainment events, and even common household activities \cite{JISE,Rajat,Jalwa,Khushbu}. The process of trying to predict the weather based on recurring meteorological and astronomical events began with early human civilizations \cite{Chauhan,IOSR,PCS}.

A vast amount of meteorological data is available even for the general public which encourages the need to monitor the weather and atmospheric changes on a daily basis \cite{IJNSA,Animesh,EPS,Kamal}. Many satellites just record atmospheric parameters and conditions \cite{IJAST,Kaur2018,CC,Jadon}. Thus this large amount of data can be used by machine learning models as an advantage over traditional techniques to save time and increase accuracy \cite{Abrahamsen,Gautam,Hybrid,OnILIS}. We used Multiple Linear Regression with the data that are given at hand to predict the temperature of a city for the upcoming days. The main focus is made on accuracy which has to be more than the traditional methods. 

\section{Related Work}
As the data is available cheaply and in vast quantity, researchers have previously tried to propose models for predicting the weather. The nature of weather is of course non-linear which makes the accuracy of prediction fairly lower as expected \cite{ICCNSJapan,HISA-SMFM,Vartika,Ankit,Preetesh}.
In paper \cite{SaxenaANN}, a review of scientific studies based on the prediction of weather using artificial neural networks is carried out. The benefits of using artificial neural networks consisted of yielding good results. Hence it can be used as an easy alternative to the traditional meteorological approach. Artificial Neural Network can be seen as capable of predicting most of the weather phenomena including rainfall, wind speed, temperature, etc. \cite{Devi} used backpropagation neural network model and tested the idea with a real-time dataset. They compared the results with the practical work of the meteorological department. The results were confirmed that the model has the potential for successful application to temperature forecasting.  

Predict and classification of thunderstorms using ANN is also done by \cite{Anad}. They designed the model to predict the occurrence of thunderstorms in two geographical regions. ANN can be used effectively for the forecasting of thunderstorms with a more than satisfactory level of accuracy. Derivation of artificial neural networks for the purpose of weather prediction for a particular location is done by \cite{Sawale}. They used a backpropagation neural network model for initial modeling. Then the results obtained by the BPN model are used to feed the Hopfield Networks. They used attributes such as wind speed, humidity, and temperature. They collected three years of data comprising of around 15000 instances.\cite{Badhiye} applied the K- Nearest Neighbor method to find the hidden pattern in a large dataset containing meteorological data of a particular area. They achieved a very high rate of accuracy but their model is unsuitable to use in remote areas.

\section{Proposed Model}
The working of the proposed model is described in this section followed by accuracy computation of the model.
We are trying to predict the mean temperature of a day using the weather data for the last 3 days and the regression technique of machine learning. We have used the Multiple Linear Regression algorithm to train our model. Also, we have applied the feature selection method to select the most decisive features out of all the features available in our initial dataset. A diagram of this model can be seen in Fig. 1. 
\begin{figure}[h]
	\centering
	\includegraphics[width=0.45\textwidth]{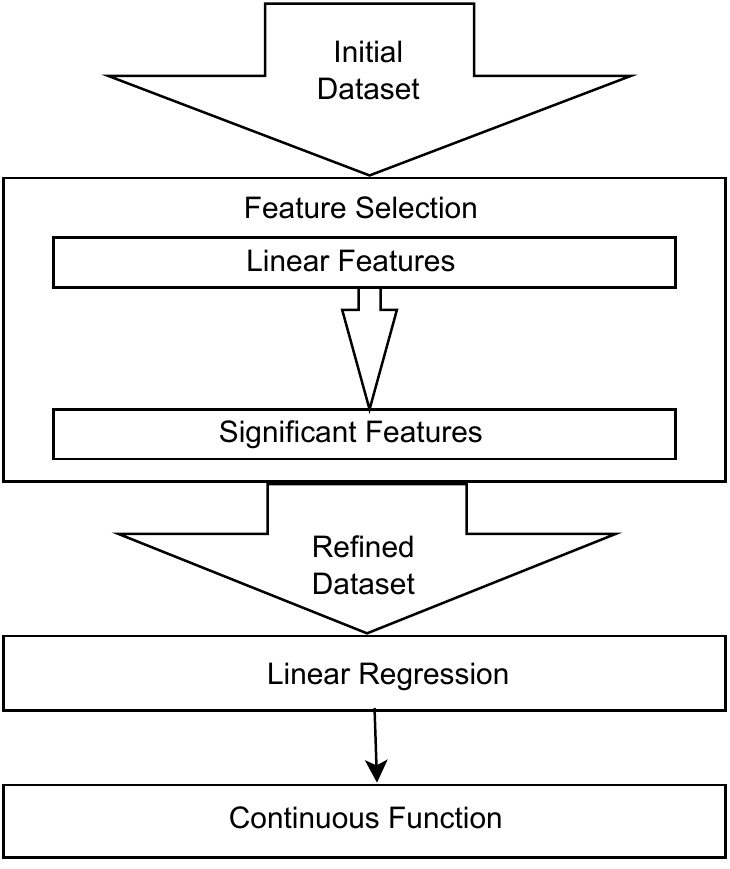}
	\caption{Block diagram of proposed model}
	\label{fig:my_label1}
\end{figure}

The initial dataset is passed to the feature selection process which does two tasks, first, it removes all the features that are not linearly dependent on the dependent variable, and then it removes all the features that are not significant. The output of this block is a refined dataset which is then passed to the next process which uses this dataset to train our machine. The output of the training process is a continuous function that is used to predict the temperature.

\subsection{Feature Selection}
An important assumption in applying multiple linear regression is that there should be a linear relationship between dependent and independent variables. To check the linear behavior between dependent and independent variables, we calculate the Pearson correlation coefficient \lq r\rq between all possible pairs of dependent and independent variables. Pearson’s correlation coefficient’s value lies between -1 and 1 inclusively, where negative value shows negative correlation and positive value shows positive correlation. For a sufficient linear relationship to exist, the value of r should be either less than -0.6 or more than 0.6. So, we discarded all the features where the absolute value of r is less than 0.6.

After this, we applied the Backward Elimination method to remove insignificant features. In Backward elimination, we use a hypothesis test. The idea is to check the effect of the removal of a feature on the model. We calculate the P-value to prove the insignificance of a feature. We first select a significance level (in our case it is 5\%), then we train our model and check the P-value of all the features. The maximum P-value found is compared with the significant level and if found greater, the corresponding feature is discarded and the process is repeated with the remaining features until all the features have a P-value lesser than the significant level. 
 
\subsection{Model Training by Multiple Linear Regression}
After feature selection is done, we obtain a refined dataset. We use this dataset to train our machine using a multiple linear regression algorithm. The dataset is split into two parts, one is used to train the model and the other is used to test the performance of the prediction. The training set stores 80\% of the data and the test set stores 20\% of the data.

The formula to predict the value in multiple linear regression model (or the regression function) is given in Eq. (1). Here, $k_0$, $k_1$....$k_n$ in  are the free parameters, y is the predicted value, and $x_1$, $x_2$…. $x_n$ are the values of the features. Once the value is predicted, we calculate the square of the difference between actual and predicted values. Then we add all the square values, and this sum is known as the error. The objective of the algorithm is to minimize this error.

\begin{equation}
 y=k_0+{k_1}*x_1+{k_2}*x_2+{k_3}*x_3.......+{k_n}*x_n  
\end{equation}

\section{Results and Analysis}  
\subsection{Experimental Setup}
For our model, the dataset that we have used is created using the Weather Underground's API web service. This dataset has 997 instances, and each instance stores the mean temperature of a day along with the weather details for the last 3 days. The weather details for the last three days include mean temperature, mean dew point temperature, mean humidity, precipitation, etc. of each day.  

\begin{figure}
	\centering
	\begin{tabular}{l c}
		&  meantempm \\
		maxdewptm\_3  & 0.829230\\
		maxtempm\_3  & 0.832974\\
		mindewptm\_3  & 0.833546\\
		meandewptm\_3  & 0.834251\\
		mimtempm\_3  & 0.836340\\
		maxdewptm\_2  & 0.839893\\
		meamdewptm\_2  & 0.848907\\
		mindewptm\_2  & 0.852760\\
		mintempm\_2  & 0.854320\\
		meamtempm\_3  & 0.855662\\
		maxtempm\_2  & 0.863906\\
		meantempm\_2  & 0.881221\\
		maxdewptm\_1  & 0.887235\\
		meandewptm\_1  & 0.896681\\
		mindewptm\_1  & 0.899000\\
		mintempm\_1  & 0.905423\\
		maxtempm\_1  & 0.923787\\
		meantempm\_1  & 0.937563\\
		meantempm  &  1.000000\\
	\end{tabular}
	\caption{Absolute values of correlation coefficient of all the possible pairs of dependent and independent variables sorted in ascending order}
	\label{fig:my_label2}
\end{figure}

\subsection{Feature Selection Results}
  To select linear features, the \lq corr\rq function of Pandas library in Python was used to calculate the correlation coefficient(r). Fig. 2 shows the value of $r$ after the removal of non-linear features. You can see in the diagram that all the values are greater than 0.6. We have used the Stats Models Library available in Python to get the P-value for all features. Fig. 3 shows the results of the final features selection. Here, it can be seen that all the P-values are less than the significant level (0.5). So, after the completion of the feature selection step, we are left with 7 features that will be used as input to multiple linear regression.
 
\begin{figure}
    \centering
    \includegraphics[width=\textwidth]{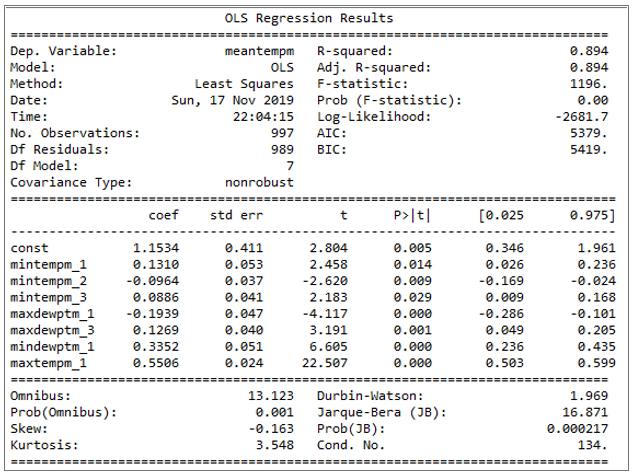}
    \caption{Summary of trained model after applying backward elimination}
    \label{fig:my_label3}
\end{figure}

\subsection{Performance Results}
We used sklearn to train our model. Sklearn is a library for python users, it provides multiple methods that implement various machine learning algorithms and other different machine learning tasks.
After training our model, we predict the temperature of test set instances, we then plot a graph using matplotlib in Python to have a graphical view of the accuracy of our model. We draw a scatter plot between predicted values and actual values. The closer the scatter plot is to line y = x, the model is more accurate. Fig. 4 shows the scatter plot of the results obtained. We can see in Fig. 3 that our scatter plot is closer to line y = x, which implies good accuracy.

\begin{figure}[h]
    \centering
    \includegraphics[width=0.6\textwidth]{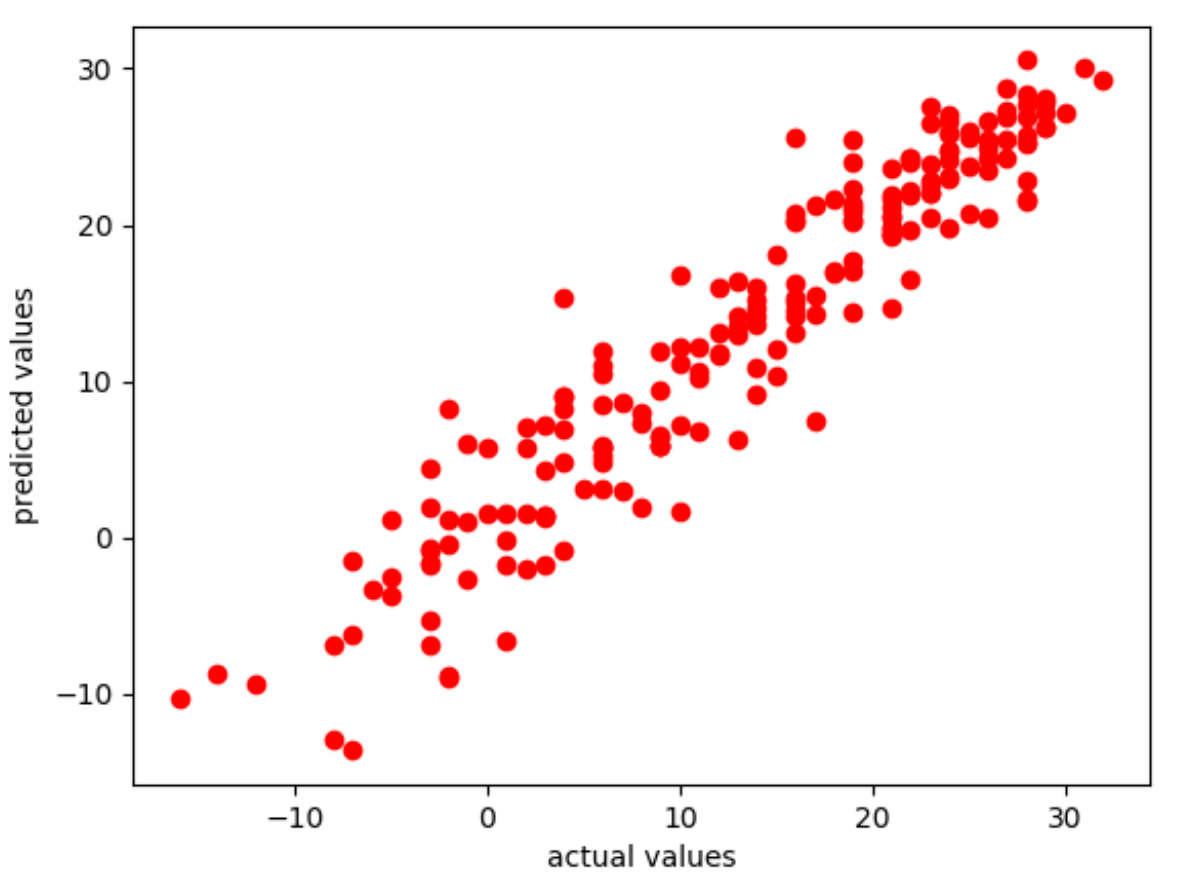}
    \caption{Scattered plot diagram between actual and predicted values}
    \label{fig:my_label4}
\end{figure}

\section{Conclusion}
A significant number of study has been done on weather forecasting in the past to benefit human society as an accurate weather prediction can contribute to saving losses including financial and human losses, and to predict day-to-day functioning. Meteorological data is available freely or at a low cost with numerous features and instances. Feature selection is important to reduce the number of redundant and unrequired features. We started with more than 30 features and they were reduced to 7. We calculated the absolute mean error on our test set. The absolute mean error thus obtained in the end is 2.8, which implies that our model can predict the mean temperature of a day with an error of 2.8 degrees Celsius, given the weather information of the last three days. The deviation of 2.8 degrees Celsius is not much so the model can be used in a simulation to get the idea of what the temperature is going to be.


\end{document}